\documentclass[10pt,twocolumn,letterpaper]{article}

\pdfoutput=1
\usepackage{cvpr}
\usepackage{times}
\usepackage{epsfig}
\usepackage{graphicx}
\usepackage{amsmath}
\usepackage{amssymb}
\usepackage{subcaption}
\usepackage{algorithm}
\usepackage{algorithmic}

\usepackage{multirow}
\usepackage{siunitx}

\usepackage[pagebackref=true,breaklinks=true,letterpaper=true,colorlinks,bookmarks=false]{hyperref}

 \cvprfinalcopy 

\def\cvprPaperID{****} 

\ifcvprfinal\fi
\begin{document}

\title{ Data-Efficient Ranking Distillation for Image Retrieval }

\author{Zakaria Laskar\\
Aalto University\\
Espoo, Finland\\
{\tt\small zakaria.laskar@aalto.fi}
\and
Juho Kannala\\
Aalto University\\
Espoo, Finland\\
{\tt\small juho.kannala@aalto.fi}
}

\maketitle

\begin{abstract}
Recent advances in deep learning has lead to rapid developments in the field of image retrieval. However, the best performing architectures incur significant computational cost. Recent approaches tackle this issue using knowledge distillation to transfer knowledge from a deeper and heavier architecture to a much smaller network. In this paper we address knowledge distillation for metric learning problems. Unlike previous approaches, our proposed method jointly addresses the following constraints i) limited queries to teacher model, ii) black box teacher model with access to the final output representation, and iii) small fraction of original training data without any ground-truth labels. In addition, the distillation method does not require the student and teacher to have same dimensionality. Addressing these constraints reduces computation requirements, dependency on large-scale training datasets and addresses practical scenarios of limited or partial access to private data such as teacher models or the corresponding training data/labels.

The key idea is to augment the original training set with additional samples by performing linear interpolation in the final output representation space. Distillation is then performed in the joint space of original and augmented teacher-student sample representations. Unlike previous mixup methods, our augmented samples are only used to generate additional training signals for the original samples and are themselves not used in the optimization process. We additionally propose algorithms to curate the idea of mixup based dataset augmentation for the problem of metric learning. Results demonstrate that our approach can match baseline models trained with full supervision. In low training sample settings, our approach outperforms the fully supervised approach on two challenging image retrieval datasets, ROxford5k and RParis6k \cite{Roxf} with the least possible teacher supervision. 

\end{abstract}

\section{Introduction}

Instance level image retrieval have dramatically improved with the advent of Convolutional Neural Networks (CNN) ~\cite{Rad_ECCV16,Gordo_ECCV16,DELF_CVPR19}. The improvement in performance is particularly driven by deeper networks such as VGG \cite{Vgg}, ResNet \cite{Resnet} family of networks. However, with the increased accuracy also comes higher inference time and computational burden at test time. There are two main ideas that have been proposed to address this challenge. One is to quantize(and/or prune) the trained bigger network to a lighter version with reduced precision and weights but with the same depth \cite{yang2019quantization}. The other direction is to transfer knowledge from the bigger network (teacher model) to a different but much smaller and lighter network (student model). In this paper, we focus on the second direction, popularly known as Knowledge Distillation (KD), although it can be applied to the former case as well.

The idea of using information from a teacher model(s) to train a student model was first proposed by Caruana \cite{Caruana_KDD06}, and, was later improved upon by Hinton \cite{Hinton_NIPSW15}. Instead of providing a one hot vector as target or ground-truth class label, KD aims to distill additional information from the teacher to the student model. Such additional knowledge is usually constituted by the output at various layers of the teacher model, e.g. logits from the layer before the softmax in the teacher constitute softer targets for the student. Traditionally proposed for classification problem, KD was later extended to the metric learning scenario ~\cite{DarkRank_AAAI,RKD_CVPR19}. However, the knowledge being distilled was addressed differently as traditional KD methods did not perform well in this setting ~\cite{DarkRank_AAAI,RKD_CVPR19}. While \cite{DarkRank_AAAI} proposed to distill the teacher ranking, \cite{RKD_CVPR19} proposed to distill the teacher distance for a given query-database sample(s). In both cases, the student model tries to learn the relation (rank/distance) between query-database samples instead of learning the exact input-output mapping. This allows the student network to maintain its own output dimensionality. We refer to these methods as Metric Knowledge Distillation (MKD) methods.

In this paper we address MKD from the perspective of data-efficiency. While existing distillation approaches ~\cite{DarkRank_AAAI,RKD_CVPR19} have addressed test time efficiency by compressing the knowledge from cumbersome models onto compact ones, they have failed to address training time efficiency. We address this issue by defining the training time efficiency as i) the number of queries to the teacher model to obtain teacher knowledge (pseudo ground-truth) in the form of final output representation or logits, and ii) the number of training samples required to distill the teacher knowledge onto the student model. In this paper we propose an MKD method under the above mentioned budget constraints while operating under the setting of black-box teacher models, preserving student-teacher dimensionality and achieving comparable performance to the no-budget scenario. Large scale datasets are costly in terms of memory (storage), computation (training) and economic (data accumulation/labelling). In addition private data such as trained teacher models or full training dataset can have limited or partial access due to privacy concerns. Our proposed method reduces the dependency on large scale datasets for learning new models while being efficient also during training.

The key ingredient in our proposed method is the idea of mixup ~\cite{InputMixup,ManifoldMixup}. Using mixup, one can augment a small original training set with large number of additional samples by convex combination of the training samples. Such idea has been recently used in \cite{KD_ALCVPR2020} to address data-efficient knowledge distillation using mixup based data augmentations. While the existing mixup based methods have addressed classification problems, we extend the idea to the problem of metric learning. In contrast to ~\cite{InputMixup,ManifoldMixup,KD_ALCVPR2020}, we perform mixup at the global image representation level. That is, each image is represented by a global representation vector obtained by spatial encoding of 2D representation maps from a CNN. Thereafter, augmented representations are obtained by linearly interpolating between representations from original samples. We then perform distillation between teacher and student models in the joint space of original and augmented global representations. In particular we train the student model to mimic the teacher ranking for each sample in the joint representation space of respective models using the recently proposed ranking loss ~\cite{Jerome_ICCV}. Representation level mixup requires orders of magnitude less queries to the teacher model compared to mixup at the input image level \cite{KD_ALCVPR2020}. In the process our proposed method still achieves comparable performance to fully supervised models trained on the full training dataset.

\section{Related Work}

We describe the image retrieval and knowledge distillation based related work in this section

\noindent \textbf{Image Retrieval} Seminal work of Sivic \etal \cite{Sivic} broadened the scope of image retrieval. The authors proposed to use SIFT \cite{SIFT} detectors and descriptors in a bag-of-features representations. To address accurate but faster large scale image the idea of large vocabularies and inverted index were extended from text search to the field of image retrieval \cite{Gordo_ECCV16_20}. An additional step of geometric verification greatly boosted the retrieval performance \cite{Gordo_ECCV16_23}. To further accelerate the image matching process for scalability on large databases, approximate methods based on spatial encoding of local descriptors were later proposed ~\cite{FV,VLAD}. 

With the success of CNN in the domain of image classification \cite{Imagenet}, attention of the image retrieval community concurrently shifted to the applicability of learnt features for this problem. The first approaches ~\cite{Sharif,Babenko} used off-the-shelf features pre-trained on ImageNet \cite{Imagenet} dataset. While some success was achieved compared to global representation based methods, the performance still lacked compared to the state-of-the-art. Later improvements focused on improving the encoding step while keeping the model parameters fixed ~\cite{Gordo_ECCV16_32,RMAC,Gordo_ECCV16_15}. 

While significant improvements were made solely based on improvements in encoding methods, the difference in data distribution between ImageNet and standard image retrieval benchmark datasets implied the features were not directly transferable. However, large scale retrieval datasets are not ubiquitous and generating ground-truth labels is a costly process. Hence, a few concurrent works ~\cite{Rad_ECCV16,Gordo_ECCV16} and \cite{netvlad} proposed fine-tuning ImageNet trained model parameters on large scale landmark datasets \cite{Datatourism}. The labels were generated using Structure-from-Motion pipeline in a self supervised manner without any human intervention. The performance of the CNN based descriptors since then have been the state-of-the-art on retrieval benchmarks ~\cite{Rad_TPAMI,DELF_CVPR19}. 

Training the CNN models requires the choice of appropriate objective functions. Ranking loss functions such as contrastive, triplet and quadruplet losses have been the popular methods for computing error function ~\cite{Gordo_ECCV16_37,quadruplet,Gordo_ECCV16}. Standard classification losses have also performed equally well \cite{GLD}. Unlike the aforementioned losses operate on a pairwise/triplets or quadruplets of samples, a more recent approach\cite{Jerome_ICCV} proposes to operate the error function on large batches of images using the proposed mean Average Precision (AP) loss. This relieves the standard process of mining hard negatives to learn robust representations.

\noindent \textbf{Knowledge Distillation} 
Knowledge distillation can be traced back to the work of Breiman \etal \cite{RKD_CVPR19_3} where the knowledge of multiple tree models were distilled onto a single model. The idea was extended to deep learning by Bucila \etal who proposed to learn a single neural network using the knowledge of an ensemble of classifiers. The term knowledge distillation (KD) was itself coined by Hinton \etal in his work \cite{Hinton_NIPSW15}. In addition to the standard supervisory signals, the student model was additionally trained to match the softmax distribution of a heavier teacher model. Since then, several works have been proposed that provide additional information apart from the softmax logits. \cite{RKD_CVPR19_47} and \cite{RKD_CVPR19_12} propose to transfer attention maps. Self-distillation where the student and the teacher share the same network architecture have also been proposed ~\cite{RKD_CVPR19_2,RKD_CVPR19_9}. 

For image retrieval and metric learning in general, Chen \etal \cite{DarkRank_AAAI} propose to transfer rank information. Similarly, Park \etal \cite{RKD_CVPR19} proposed to distill distance information between teacher and student models. Both these methods show improvements over standard KD approaches.

\noindent \textbf{Mixup.}Mixup based  regularizers were first proposed by ~\cite{InputMixup,tokozume_cvpr2018}. Later, mixup based interpolation was extended to hidden representations of a CNN by several works ~\cite{ManifoldMixup,ZhaoCho}. Recently, \cite{KD_ALCVPR2020} proposed mixup based augmentation for data-efficient knowledge distillation. For each mixed sample, the above approaches require a new feed-forward pass through the CNN. Instead our approach performs mixing of the global vector representations requiring just a single feed-forward pass. Mixed samples can be obtained by simply interpolating between original global representations.

\section{Proposed Method}
In this section we propose an algorithm to train a compact student model, $S$ by distilling the knowledge from a cumbersome teacher model, $T$. The key idea of knowledge distillation in classification problem is to use soft labels from $T$ as targets in addition to ground-truth hard labels. Soft labels encode semantic similarity by providing inter class similarity information.  However, we consider a general scenario where ground-truth labels are not known apriori. In addition metric learning involves optimizing the representation space directly without explicit label prediction as in classification problems. Thus it is not clear how to generate and incorporate teacher soft labels for unsupervised knowledge distillation in the domain of metric learning problems.

First we present preliminaries followed by the data augmentation algorithm to address low training sample complexity. This is followed by teacher label generation and computing the ranking loss to train the student model. Finally, we present the algorithm combining the above steps in a single framework.

\label{sec:algo}
\subsection{Preliminaries}
Given a batch of images, $B = \{I_1,..I_j..I_B\}$, we obtain teacher and student $l2$ normalized output representations, $f_B^T = \{f_b^T\}_{b=1}^{|B|} \in \mathbb{R}^{N_T \times |B|}$ and $f_B^S = \{f_b^S\}_{b=1}^{|B|} \in \mathbb{R}^{N_S \times |B|}$, where $f_b^T = e^T(T(I_b)) \in \mathbb{R}^{N_T}$ , $f_b^S = e^S(S(I_b))\in \mathbb{R}^{N_S}$. $T(.)$ and $S(.)$ are teacher and student convolutional neural networks respectively, with $N_T, N_S$ being their respective final output dimensionality. As database size in image retrieval problems tends towards millions, it is common practise to store global representations per image by encoding the 3D representation map from the CNNs into 1D vectors. Popular encoding methods from literature include GeM \cite{Rad_TPAMI}, MAC \cite{Rad_ECCV16}, RMAC \cite{RMAC}. As we consider the teacher model as black-box with access to the final vector encoded global representation, we represent the student and teacher encoding functions separately with $e^T$ and $e^S$ respectively. These encoding functions can represent any of the above mentioned encoding methods. As the global representations are $l2$ normalized, a simple dot product is used to compute similarity values.


\subsection{Database Augmentation}
Acquiring large training datasets and labelling the ground-truth incurs large computational resources, huge memory footprint and high economic costs. We address this using knowledge distillation by replacing large datasets with models trained on them and a small amount of the original training samples. This also addresses practical scenarios where teacher models have limited access rights, or the whole training set is not made public. Furthermore, extracting representations for the whole dataset using both the teacher and student model is inefficient as it leads to increased training costs.

Given a small amount of training samples, $D$, we extract both teacher and student global representations, $f^T$ and $f^S$ for a given batch, $B$. We augment the representations from each batch using mixup \cite{InputMixup},\cite{ManifoldMixup}. These works perform mixup at the local level, while we perform mixup at the global representation level. In particular, given representations for images, $I_i, I_j \in B$, we perform representation mixup as follows:

\begin{equation}
    f_{ij} = \lambda f_i + (1 - \lambda) f_j,
\label{eq:mix}
\end{equation}

\noindent where $\lambda \sim beta(\alpha,\alpha) $ is the mixing coefficient. Instead of sampling $\lambda$ per training sample, we only sample a single value of $\lambda$ per batch. The mixed representations are further $l2$ normalized.

There are several benefits to performing mixup at the global representation level. It is to be noted that since we consider black-box teacher models, we can only consider \textit{InputMixup} \cite{InputMixup} and not \textit{ManifoldMixup} \cite{ManifoldMixup} as the later requires access to intermediate representation maps. However, \textit{InputMixup} which performs mixup at the input image level, requires a new feed-forward pass of the mixed input image through the network to obtain representations. This increases the number of queries to the teacher model to be much more than $|D|$. The same applies for the student network, which in total significantly increase the training cost. The same costs applies for \textit{ManifoldMixup} considering white-box teacher models. In contrast, global representation mixup only requires at most $|D|$ queries to the teacher model, and large amount additional representations can be obtained at a marginal overhead cost.
To give the reader an estimate, given a batch of $B$ = 1000 images, mixing each image with $R$ = 10 other images from the batch will result in 10000 samples. For \textit{InputMixup} this will require 10000 $\times \textit{ep}$ queries to the teacher model where $\textit{ep}$ is the number of training epochs. Our approach will only require 1000 queries.

\subsection{Label Generation}
Previously, mixup has been addressed in classification domains ~\cite{InputMixup,ManifoldMixup,CutMix}. In such settings, the label of the mixed sample is obtained by linear interpolation of respective labels of the original samples by $\lambda$. In this section, we show how to generate labels for student model using global teacher representations.

Let the joint set of representations be $F^T = f_B^T \bigcup f_{B'}^T$, $F^S = f_B^S \bigcup f_{B'}^S$, where $f_{B'}^T$,$f_{B'}^S$ are the augmented teacher and student representations. Let $Z = \{1,2,...,(|B| + |B'|)\}$ be the joint sample index set. We are interested in computing a binary label matrix $Y \in \mathbb{R}^{|Z\times|Z|}$, where each row $Y_q \in \mathbb{R}^{1 \times |Z|}$ represents the label vector corresponding to the $q^{th}$ representation. Before we explain how the binary values are computed for $Y_q$, we first formally define a positive index set $P_q \subseteq Z$ such that $\forall z \in Z$


\begin{equation}
Y_q(z)= 
\begin{cases}
    1, & \text{if}\ z \in P_q  \\
    0, & \text{otherwise}
\end{cases}
\label{eq:Y_q}
\end{equation}

\noindent The matrix $Y$ is symmetric ($Y = Y^\intercal$) i.e. $Y_q(z) = Y_z(q)$. The binary label $1$ signifies the corresponding representations, $f_q^T, f_z^T$ are similar. Consequently, the corresponding student representations, $f_q^S, f_z^S$ are trained to be similar. The measure of similarity is defined next where we present methods to compute the elements of $P_q$.

\noindent \textbf{Similarity Labelling (SL)}.
The first measure of similarity is based on cosine similarity in the representation space. We first compute the teacher and student similarity matrices, $S^T = (F^T)^\intercal(F^T)$, $S^S = (F^S)^\intercal(F^S) \in \mathbb{R}^{|Z|\times|Z|}$. The positive set $P_q$ constitutes the Euclidean Nearest Neighbors (ENN) and is computed as $P_q = \{z \hspace{.2mm} | \hspace{.2mm} S_q^T(z) > \tau\} \forall z \in Z$, where $\tau \in [0,1]$ is a similarity threshold. We call this similarity based labelling Similarity Labelling (SL). If $\tau$ is too high, $P_q$ will only contain near duplicate representations, while keeping it too low will include too many false positives. We experiment with different values of $\tau$ and found that optimal performance is achieved with moderate values of $\tau$ (c.f. Sec. \ref{sec:hyp_abl}). 

\noindent \textbf{Mixup Labelling.} As observed the positive set, $P_q$ under similarity labelling is constituted by the ENN. Thus for representations falling in low density regions, the positive sets will be empty or have low cardinality. Empty positive sets means zero loss and thus no gradient to train the model. This becomes an issue if most of the samples fall in such low density regions (c.f. Sec. \ref{sec:tr_abl}). To address this issue we introduce mixup labelling (ML) based on the following assumption: the global representations contain semantic concepts, so mixed representation will be closer to the positive sets of representations being mixed, than the other representations. Formally, if representations, $f_k^T, f_r^T$ are mixed resulting in the corresponding mixed representation, $f_{kr}^T$ indexed at $kr \in Z$, and the corresponding positive sets obtained from similarity labelling be $P_k, P_r$ and $P_{kr}$ respectively. Then $P_{kr} = P_{kr} \cup P_k \cup P_r$. Thereafter, using Eq. \ref{eq:Y_q} the label matrix, $Y$ is computed while maintaining matrix symmetry.



\subsection{Loss Function}

In this section we show how to compute the loss given the student similarity matrix, $S^S$ and teacher label matrix, $Y$. To realize this, we use the listwise loss, known  as  Average Precision Loss (AP) \cite{Jerome_ICCV}. The loss maximizes the distance between histogram of positive and negative similarity scores. For brevity we elaborate the loss function below:

The similarity interval $S_q^S = [0,1]^{1 \times |Z|}$ is divided into $C-1$ bins of width $\Delta c = \frac{2}{C-1} $. Let $c_b = 1 - (b-1)\Delta c$, $b$=1...C represent the center of the $b^{th}$ bin. Average Precision is generally computed at each rank, $r = 1 ... |Z|$. However, as rank assignment is non-differentiable, the images are instead assigned to bins using the soft bin assignment as follows:

\begin{equation}
p(S_q^S(i),b) = \max \left( 1-\frac{|S_q^S(i)-c_b|}{\Delta}, 0\right).\label{eq:binning}
\end{equation}
Here, $p$ represents the probability that the $i^{th}$ image occupies the $b^{th}$ bin conditioned on its similarity to the query, $q$ based on $S_q^S$. The AP is then computed in each bin as follows:

\begingroup
\vspace{-2mm}
\addtolength{\jot}{0.5em}
\begin{align}
{Pr} (S_q^S,Y_q,b)&= \dfrac{\sum_{b'=1}^b p(S_q^S,b')^\top Y_q }{\sum_{b'=1}^b p(S_q^S,b')^\top \text{\textbf{1}}}, \\
\Delta {Rc} (S_q^S,Y_q,b)&= \dfrac{p(S_q^S,b)^\top Y_q}{N_q},
\end{align}
\endgroup

\noindent The AP loss for each $q$ is computed as :
\begin{equation}
AP(q) = 1-\sum_{b=1}^{C}{Pr}(S_{q}^S,Y_{q},b)\ \Delta {Rc}(S_{q}^S,Y_{q},b).\label{eq:apquant}
\end{equation}

The final loss function to be optimized is defined as :

\begin{equation}
L = \frac{1}{|Z|}\sum_{q=1}^{|Z|}{AP}(q)
\end{equation}

\subsection{Algorithm}

\begin{algorithm}[t]
\caption{Rank Distillation }
\hspace*{0.02in}
\label{algo:loss}

{\bf REQUIRE:} Teacher \& Student representation $\{f_i^T\}_{i=1}^B, \{f_i^S\}_{i=1}^B$

{\bf REQUIRE:}Labelling functions: $SL(.), ML(.)$

{\bf REQUIRE:}Loss function: $AP(.)$

{\bf REQUIRE:} $\tau, R, \lambda$

{\textbf{OUTPUT:}}  Loss value 

\begin{algorithmic}[1] 
\STATE{Initialize $L = \{\}$ $\triangleright$ store loss values.}
\FOR{ $r = 1, 2..., R$}
        \STATE{Initialize $F^T$ = \{\}, $F^S$ = \{\} $\triangleright$ store original \& augmented samples.}
        \STATE{Initialize $MIX = \{\}$ $\triangleright$ store indices of mixing \& mixed samples.}
        \STATE{$\{F_i^T\}_{i=1}^{|B|} \leftarrow \{f_i^T\}_{i=1}^{|B|}$, $\{F_i^S\}_{i=1}^{|B|} \leftarrow \{f_i^S\}_{i=1}^{|B|}$}
        \FOR{$k = 1, 2..., |B|$}
            \STATE{Sample index $r_k$ from range $(1,|B|)$}
            \STATE{$F_{|B|+K}^T \leftarrow \lambda f_k^T + (1-\lambda)f_r^T $}
            \STATE{$F_{|B|+K}^S \leftarrow \lambda f_k^S + (1-\lambda)f_r^S $}
            \STATE{$F_{|B|+K}^S$.requires\_grad = False} $\triangleright$ Pytorch format to remove variable from computational graph.
            \STATE{$MIX$.store($k,r,|B|+k$) $\triangleright$ Store mixing information.}
        \ENDFOR
        \STATE{$S^T \leftarrow (F^T)^\intercal(F^T)$, $S^S \leftarrow (F^S)^\intercal(F^S)$ $\triangleright$ Compute teacher \& student similarity matrix }
        \STATE{$P \leftarrow SL(S^T, \tau)$. $\triangleright$ Initial positive set based on SL. }
        \STATE{$P \leftarrow ML(P, MIX)$ $\triangleright$ Final positive set based on ML.}
        \STATE{Compute $Y$ using $P \triangleright$ Eq. \ref{eq:Y_q}}
        \STATE{$L_r \leftarrow AP(Y, S^S) \triangleright$ Compute Average Precision Loss}
\ENDFOR
\STATE{$L \leftarrow 1/R{\sum_{r=1}^R L_r} $ $\triangleright$ Total loss }
\RETURN{$L$}
\end{algorithmic}
\end{algorithm}

Given the training set, $D$ we first extract all the teacher representations, $f_D^T = \{f_i^T\}_{i=1}^{|D|}$. Thereafter for each epoch, $ep$, we sample a batch of $B$ images from $D$. We then extract the student representations, $f_B^S = \{f_b^S\}_{b=1}^{|B|}$ and from $f_D^T$ obtain the teacher representations $f_B^T = \{f_b^T\}_{b=1}^{|B|}$. We now proceed to compute the loss as described in Algorithm \ref{algo:loss}. 

First, we introduce the mixing iterator, $R$. In each iteration, $r = 1,2,...R$ we iterate over the following steps: 1)Mix each student and teacher representation, $f_k^T, f_k^S, k=1,2..|B|$ with the representation, $f_r^T \in f_B^T, f_{r_k}^S \in f_B^S$ of a random sample $r_k$. The mixed representations are concatenated with the original representations resulting in the joint representation set, $F^T \in \mathbb{R}^{N_T \times 2|B|}$, $F^S \in \mathbb{R}^{N_S \times 2|B|}$, where the first $N \times |B|$ are the original representations while the bottom $N \times |B|$ are mixed representations respectively. 2) Simultaneously with the previous step we store the index information of mixing samples ($k,r_k$) and the mixed sample ($|B|+k$), $\forall k=1...|B|$ in the variable, $MIX$. 3) Given $F^T, F^S$ we compute the teacher and student similarity matrices, $S^T, S^S$. 4) Next we proceed to label generation. Using the similarity threshold, $\tau$ and $S^T$ we first compute the positive set, $P = \{P_k\}_{k=1}^{2|B|}$ based on similarity labelling (SL). Thereafter, using the mixing information in $MIX$ and $P$ we perform mixup labelling (ML) to compute the final positive set, $P$. The label matrix, $Y \in \mathbb{R}^{2|B| \times 2|B|}$ is formed using $P$. 5) Finally, we compute the Average Precision (AP) loss, $L_r$ using teacher label matrix, $Y$ and student similarity matrix, $S^S$. After $r$ iterations we have $R$ loss values, $\{L_r\}_{r=1}^R$ which are then averaged followed by back-propagation. It is to be noted that the mixed representations are not used to back-propagate gradients (line 10 in Algorithm \ref{algo:loss}). 

We now explain the rationale behind introducing $R$. Under the current setting, the number of mixed representations used to compute the final loss is $|B|R$.  If these mixed representations were jointly used in computing the final loss the size of the similarity and label matrices will be ($(|B| + |B|R)^2$). For values of $|B|$=1000, $R$=10 used in this work, the size will be $\sim 10000^2$ . Loss and gradient computation becomes considerably slow under this setting. Instead, by dividing the loss computation into $R$ steps, we are still able to leverage $|B|R$ mixed representations, while the final similarity matrices are of size $(2|B|)^2 \sim 2000^2 $. This leads to comparable performance while increasing training efficiency.

\section{Implementation details}
\noindent \textbf{Training dataset.} We use the training dataset used in \cite{Rad_ECCV16}. The dataset was initially introduced in \cite{Datatourism} and consists of 7.4 million internet photo collections of popular landmarks around the world. The images are passed through an SfM pipeline \cite{Rad_ECCV16} to create clusters of images (class labels). This process results in 163k images clustered into about 700 classes. Training dataset consists of randomly selected 550 classes containing 133k images. We refer to this dataset as \textit{SfMFr}

\vspace{2mm}
\label{sec:impl}
\noindent \textbf{Network Training.} 
We used the publicly available trained Resnet101 models by Radenovic \etal  \cite{Rad_TPAMI} trained on \textit{SfMFr} as teacher models, $T$. MobileNetV2 (MVNetV2) and Resnet34 pre-trained on ImageNet \cite{Imagenet} are used as the student models S1 and S2 repsectively. We randomly sample $D$ = 4000 images from \textit{SfMFr}. The network is trained with a batch size, $B$ = 1000 which are randomly sampled from the training set, $D$. We used Adam \cite{adam} optimizer with an initial learning rate of $l_0$ = \num{1e-4}, exponential decay $l_0\exp(-0.01 \times ep)$ every epoch, $ep$. Weight decay was set to \num{1e-6}. Images were rescaled to 362 pixels on the longest side while maintaining the aspect ratio. Training is done for 30 epochs on GeForce RTX GPU with 11GB memory.
We use generalized mean pooling (GeM) \cite{Rad_TPAMI} to obtain global representations for each image. The global descriptors are subsequently $l2$ normalized.

We list the hyper-parameters associated with our algorithm are $\tau$ = 0.75, $R$ = 10. We use the same hyper-parameter settings for both the student networks, S1 and S2.

\noindent \textbf{Baselines}. We train MVNetV2, Resnet34 using contrastive loss (CL) and Average Precision (AP) loss on \textit{SfMFr} dataset.  For CL, we mine hard negatives every epoch from a random pool of 22K images, and keep top 5 negatives. Margin is set to 0.65. Batch size are 5 and 4000 for CL and AP respectively. The learning rate was set to \num{5e-7} for CL and \num{1e-4} for AP.

\begin{table}[h]
\newcolumntype{L}[1]{>{\raggedright\let\newline\\\arraybackslash\hspace{0pt}}m{#1}}
\newcolumntype{C}[1]{>{\centering\let\newline\\\arraybackslash\hspace{0pt}}m{#1}}
\newcolumntype{R}[1]{>{\raggedleft\let\newline\\\arraybackslash\hspace{0pt}}m{#1}}
\def\arraystretch{1.1}
\newcommand\cw{0.7cm}
\footnotesize
\begin{center}
{%
\setlength{\tabcolsep}{0.2mm}
\begin{tabular}{|@{~}L{2.7cm}|@{~}L{2.0cm}|@{~}L{2.0cm}|}
    \hline
    \multirow{1}{*}{Method} & \multirow{1}{*}{$\#$Param} & \multirow{1}{*}{Time/Image}\\
    \hline
    ResNet101 & 42M & 60ms \\
    ResNet34 & 21M & 20ms \\
    MVNetV2 & 1.8M & 10ms \\
    \hline
\end{tabular}
\caption{Different networks with the number of respective parameters and time taken to process 1 image in multi-scale mode.}
\label{tab:param}
}
\end{center}
\end{table}

\noindent \textbf{Test dataset} We evaluate our approaches on standard image retrieval benchmark datasets, Oxford5k (Oxf) \cite{Oxf},Paris6k (Par) \cite{Par}, ROxford5k (ROxf) \cite{Roxf}, and RParis (RPar) \cite{Roxf} datasets. The evaluation metric is mean Average Precision (mAP). The test sets consists of 55 queries and several thousand database images (Oxf:5k, Par:6k), while their revisited counterparts have 70 queries each with 4k and 6k database images respectively. The revisited datasets also have 3 splits : Easy (E), Medium (M), and Hard (H) defining the difficulty level of retrieving the corresponding database images in the set . The queries are annotated with a bounding box specifying the landmark of interest. Similar to prior works, we crop the query images with bounding box. 

During evaluation, we extract multi-scale global representations with the scales: 1, 1/$\sqrt{2}$, and 1/2. The resulting descriptors are combined using GeM pooling. The resulting vector is $l2$ normalized. Furthermore, due to low sample complexity we do not use any validation data during training. Instead, we perform weight averaging \cite{wavg} to combine model performances from different epochs. In particular, the final student network used for evaluation is obtained by averaging the weights of the trained student models from the 20$^{th}$ and 30$^{th}$ epoch.

The number of parameters and average multi-scale inference time during evaluation are presented in Tab. \ref{tab:param} for teacher and student models.

\begin{table*}[h!]
\newcolumntype{L}[1]{>{\raggedright\let\newline\\\arraybackslash\hspace{0pt}}m{#1}}
\newcolumntype{C}[1]{>{\centering\let\newline\\\arraybackslash\hspace{0pt}}m{#1}}
\newcolumntype{R}[1]{>{\raggedleft\let\newline\\\arraybackslash\hspace{0pt}}m{#1}}
\def\arraystretch{1.1}
\newcommand\cw{0.7cm}
\footnotesize

\begin{center}
{%
\setlength{\tabcolsep}{0.2mm}
\begin{tabular}{|@{~}L{3.2cm}|@{~}L{1.0cm}|C{\cw}|C{\cw}|C{\cw}|C{\cw}|C{\cw}|C{\cw}|C{\cw}|C{\cw}|}
    \hline
    \multirow{2}{*}{Method} & \multirow{2}{*}{Samples} & \multirow{2}{*}{Oxf} & \multicolumn{3}{c|}{ROxf} & \multirow{2}{*}{Par} & \multicolumn{3}{c|}{RPar} \\
    \cline{4-6}\cline{8-10}
    & & & E & M & H & & E & M & H \\
    \hline\hline
    
    Resnet101 (T,CL) & \hspace{2mm}120k & 81.2 & 73.8 & 55.8 & 27.4 & 87.8 & 86.5 & 70.0 & 44.8 \\
    \hline 
    \multicolumn{10}{|c|}{\b{Compact student networks}} \\
    \hline
    MVnetV2 (CL) & \hspace{2mm}120k & 74.5 & 66.5 & 48.9 & 20.8 & 85.7 & 84.6 & 66.2 & 39.0 \\
    MVnetV2 (AP) & \hspace{2mm}120k & 74.2 & 67.2 & 51.0 & 24.3 & 85.0 & 83.7 & 65.6 & 39.5 \\
    MVNetV2 (S1, no-aug) & \hspace{3.5mm}4k & 76.1 & 67.0 & 51.0 & 25.8 & 84.6 & 84.0 & 66.1 & 40.4 \\
    
    MVnetV2 (S1) & \hspace{3.5mm}4k & 78.7 & 70.8 & 53.8 & 26.9 & 84.0 & 82.1 & 65.0 & 39.4 \\
    \hline
    ResNet34 (CL) & \hspace{2mm}120k & 77.9 & 70.7 & 51.9 & 23.1 & 86.5 & 85.9 & 69.5 & 44.0 \\
   
    ResNet34 (AP) & \hspace{2mm}120k & 79.6 & 70.5 & 53.3 & 24.9 & 86.3 & 86.4 & 69.2 & 43.1 \\
    
    ResNet34 (S2,no-aug) & \hspace{3.5mm}4k & 77.3 & 70.7 & 50.8 & 22.5 & 84.9 & 83.8 & 68.0 & 42.8 \\
    
    ResNet34 (S2) & \hspace{3.5mm}4k & 78.1 & 74.1 & 55.1 & 25.9 & 85.4 & 84.2 & 68.6 & 43.8 \\
    
    \hline
    
\end{tabular}
\caption{Performance comparison of compact student networks MobileNetV2 (S1,MVnetV2) and Resnet34 (S2,ResNet34) trained using our method, and baseline methods : without augmentation (no -aug), Average Precision (AP), contrastive loss (CL). Evaluation is done on image retrieval datasets : Oxford(Oxf), Paris(Par), ROxford(ROxf), RParis(RPar). The revisited datasets, ROxf, RPar are evaluated using Easy (E), Medium (M) and Hard (H) splits. Evaluation metric is mAP. T/S denotes the teacher /student role of the model. Our method does not require training labels.}
\label{tab:results_S_T}
}
\end{center}
\end{table*}


\section{Results}

In this section we compare our proposed algorithms on the standard retrieval datasets. In addition we also compare with baseline methods and perform detailed ablation study.

\noindent \textbf{Baseline comparison.} We compare the performance of the student models, S1: MVNetV2, S2: ResNet34 trained using our proposed method with the teacher model, T: ResNet101 and the same student models trained without the proposed augmentation. In addition we also consider student models trained with ground-truth labels on the full dataset with loss functions: contrastive loss (CL) and average precision (AP). Results are presented in Tab. \ref{tab:results_S_T}. Results show that student models using our proposed method are able to match the performance of the supervised counterparts. It is to be noted that our method was only trained on 4k images. Compared to it, the supervised models based on CL and AP losses were trained using the full dataset of 120k images. Among the student models, ResNet34 outperforms MVNetV2 by 2-3\% on ROxford and RParis datasets. This can be attributed to the higher capacity of ResNet34 model (c.f. Tab. \ref{tab:param}).

Furthermore, student models trained without the proposed global representation augmentation performs poorly compared to our proposed method with augmentation. Baseline student models trained only on $D=4k$ dataset with AP loss and full ground-truth label supervision has similar performance to the no-augmentations setting. The decrease in performance in low sample setting can be attributed to the fact that in a randomly selected training set $D$, large number of samples, $q \in D$ have empty or very small sized positive set $P_q$. Thus, without any positives there is no error signal that can lead to learning representations from these images. Augmentation addresses this issue by generating positives from the mixed samples.

 \begin{table*}[h!]
\newcolumntype{L}[1]{>{\raggedright\let\newline\\\arraybackslash\hspace{0pt}}m{#1}}
\newcolumntype{C}[1]{>{\centering\let\newline\\\arraybackslash\hspace{0pt}}m{#1}}
\newcolumntype{R}[1]{>{\raggedleft\let\newline\\\arraybackslash\hspace{0pt}}m{#1}}
\def\arraystretch{1.1}
\newcommand\cw{0.7cm}
\footnotesize

\begin{center}
{%
\setlength{\tabcolsep}{0.2mm}
\begin{tabular}{|@{~}L{3.2cm}|@{~}L{1.0cm}|@{~}L{1.0cm}|C{\cw}|C{\cw}|C{\cw}|C{\cw}|C{\cw}|C{\cw}|C{\cw}|C{\cw}|}
    \hline
    \multirow{3}{*}{Method} & \multirow{3}{*}{Nw} & \multirow{3}{*}{\hspace{2mm} Dim} & \multicolumn{4}{c|}{ROxf} & \multicolumn{4}{c|}{RPar} \\
    \cline{4-11}
    & & & \multicolumn{2}{c|}{M} & \multicolumn{2}{c|}{H} & \multicolumn{2}{c|}{M} & \multicolumn{2}{c|}{H} \\
    \cline{4-11}
    & & & \tiny mAP & \tiny mP@10 & \tiny mAP & \tiny mP@10 & \tiny mAP & \tiny mP@10 & \tiny mAP & \tiny mP@10 \\
    \hline
    GeM \cite{Rad_TPAMI} & R101 & 2048 & 64.7 & 84.7 & 38.5 & 53.0  & 76.9 & 98.1  & 55.4 &  89.1 \\
    AP \cite{Jerome_ICCV} & R101 & 2048 & 67.5 & - & 42.8 & - & 80.1 & - & 60.5 & - \\
    \hline
    GeM \cite{Rad_TPAMI} & V & 512 & 60.9 & 82.7 & 32.9 & 51.0 & 69.3 & 97.9 & 44.2 & 83.7 \\
    NetVLAD \cite{netvlad} & V & 512 & 37.1 & 56.5 & 13.8 & 23.3 & 59.8 &  94.0 & 35.0 & 73.7 \\
    \hline
    Ours & R34 & 512 & 55.4 & 79.9 & 29.1 & 46.3 & 68.7 & 96.6 & 43.7 & 83.4 \\
    Ours & M & 320 & 51.1 & 74.0 & 24.9 & 38.6 & 67.3 & 96.1 & 41.1 & 80.0 \\
    
    \hline
\end{tabular}
\caption{mAP performance on ROxford (ROxf) and RParis (RPar) datasets. We present alongside each method, the model architecture (R:ResNet101, V:VGG16,A:AlexNet,R34:ResNet34 and M:MobileNetV2). In addition we also show the dimension of the final global representation from each model. It is to be noted that our proposed method only requires a fraction of the full supervised dataset and a trained teacher.}
\label{tab:results_sota}
}
\end{center}
\end{table*}

\noindent{\textbf{State-of-the-art.}} State-of-the-art methods are compared in Tab. \ref{tab:results_sota}. Whitening is a standard post-processing step in all standard image retrieval methods as it reduces the impact of correlated features. While some \cite{Jerome_ICCV} use unsupervised whitening based on PCA, others \cite{Rad_TPAMI} use supervised whitening. We use PCA based whitening. In particular we use the square rooted PCA \cite{Jegou_ECCV12}. Similar to traditional practices, we learn PCA on Paris6k for evaluating the network on Oxford5k and vice-versa. In addition to mAP, we also report mean precision @10 (mP@10). In RPar PCA does not bring any improvement. However, in ROxf, PCA brings significant improvement both in terms of mAP : ResNet34 (ROxf,M : 55.1 $\rightarrow$ 55.4, ROxf,H : 25.9 $\rightarrow$ 29.1), and mP@10 : ResNet34 (ROxf,M : 76.6 $\rightarrow$ 79.9, ROxf,H : 39.8 $\rightarrow$ 46.3). However, there is an increase in the performance gap with the teacher model (GeM \cite{Rad_TPAMI}). The performance difference can be attributed to the supervision in the whitening process. Compared to supervised models with similar dimensionality such as (GeM \cite{Rad_TPAMI}, V) that uses a VGG16 architecture, the performance gap is much smaller. Overall, the difference in student performance is well compensated by the reduced number of parameters and computation time for processing a single image in current multi-scale mode as shown in Tab. \ref{tab:param}.

\section{Hyper-parameter Ablation}
\label{sec:hyp_abl}
 In this section, we analyze and present detailed analysis on the impact of different hyper-parameters in retrieval performance. This is done by varying the concerned hyper-parameter while keeping the rest same as detailed in Sec. \ref{sec:impl}. 
 
 First we analyze the image retrieval performance by varying the size of the training set, $D$. In Tab. \ref{tab:D_abl} we observe that our method consistently outperforms the baseline methods trained without the representation augmentations. 
 
 \begin{table}[t!]
\newcolumntype{L}[1]{>{\raggedright\let\newline\\\arraybackslash\hspace{0pt}}m{#1}}
\newcolumntype{C}[1]{>{\centering\let\newline\\\arraybackslash\hspace{0pt}}m{#1}}
\newcolumntype{R}[1]{>{\raggedleft\let\newline\\\arraybackslash\hspace{0pt}}m{#1}}
\def\arraystretch{1.1}
\newcommand\cw{0.7cm}
\footnotesize

\begin{center}
{%
\setlength{\tabcolsep}{0.2mm}
\begin{tabular}{|@{~}L{3.5cm}|C{\cw}|C{\cw}|C{\cw}|C{\cw}|C{\cw}|}
    \hline
    \multirow{2}{*}{Method}  & 
    \multirow{2}{*}{$|D|$} &
    \multicolumn{2}{c|}{ROxf} &  \multicolumn{2}{c|}{RPar} \\
    \cline{3-4}\cline{5-6}
    & & M & H & M & H \\
    \hline\hline
    
    ResNet34  & 2000 &  54.3 & 25.3 &  66.4 & 40.5 \\
    
    ResNet34  & 4000 & 55.1 & 25.9 & 68.6 & 43.8 \\
    
    ResNet34  & 8000 & 55.4 & 26.3 & 67.6 & 42.1 \\
   
    \hline
    
    ResNet34 (no-aug) & 2000 &  49.1 & 20. &  66.4 & 40.0 \\
    
    ResNet34 (no-aug) & 4000 & 50.8 & 22.5 & 68.0 & 42.8 \\
    
    ResNet34 (no-aug) & 8000 & 53.2 & 25.2 & 67.6 & 41.5 \\
    
    \hline
    
\end{tabular}
\caption{Performance comparison ROxf and RPar using mAP metric under different training budget given by the size of training set, $|D|$. Results demonstrate our proposed method (Row:1-3) based on representation augmentation consistently outperforms baseline cases (Row:4-6) without augmentation.}
\label{tab:D_abl}
}
\end{center}
\end{table}

\begin{figure*}[h]
\begin{subfigure}{.5\textwidth}
  \centering
  \includegraphics[width=.8\linewidth]{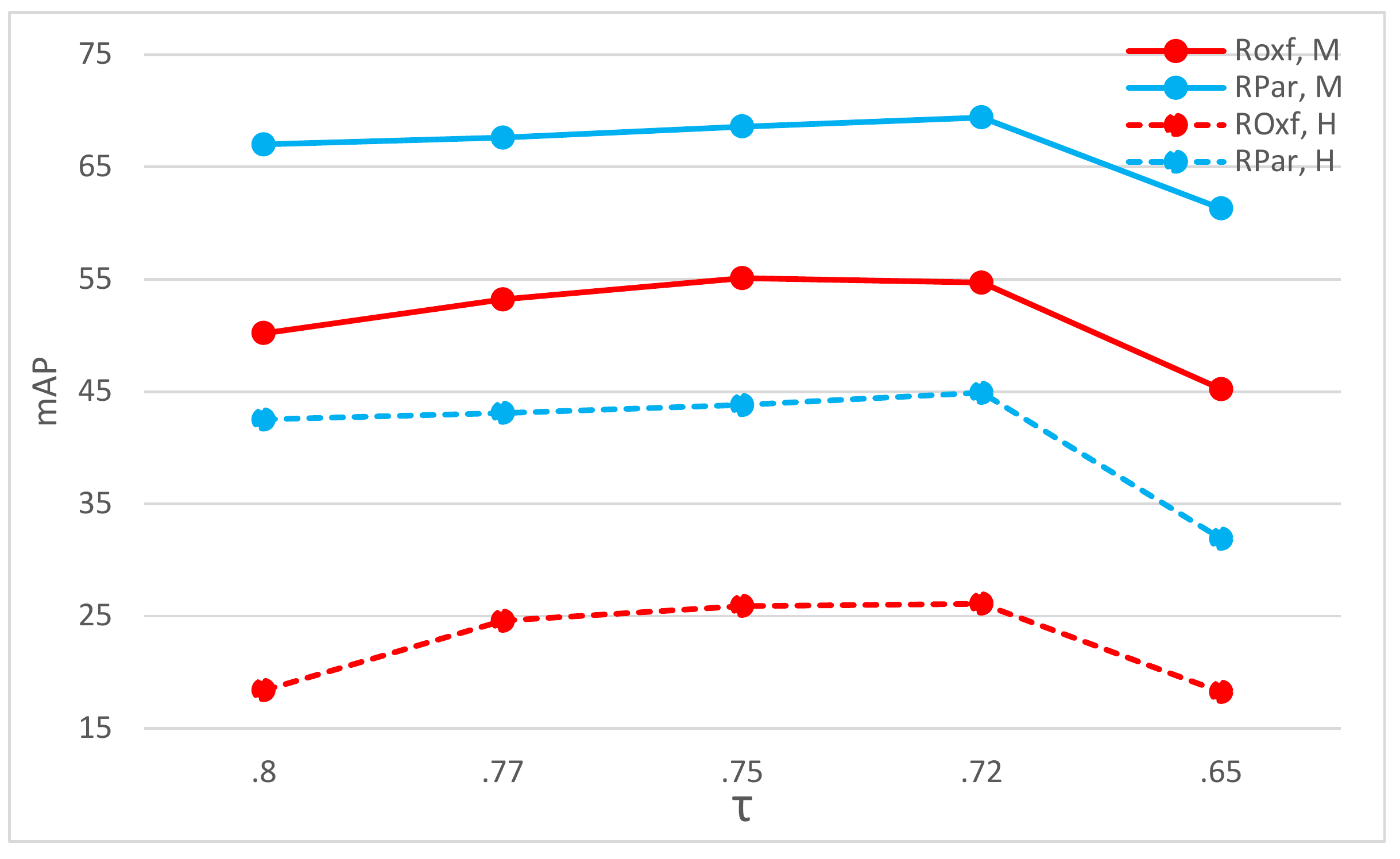}
  \caption{1a}
  \label{fig:sfig1}
\end{subfigure}%
\begin{subfigure}{.5\textwidth}
  \centering
  \includegraphics[width=.8\linewidth]{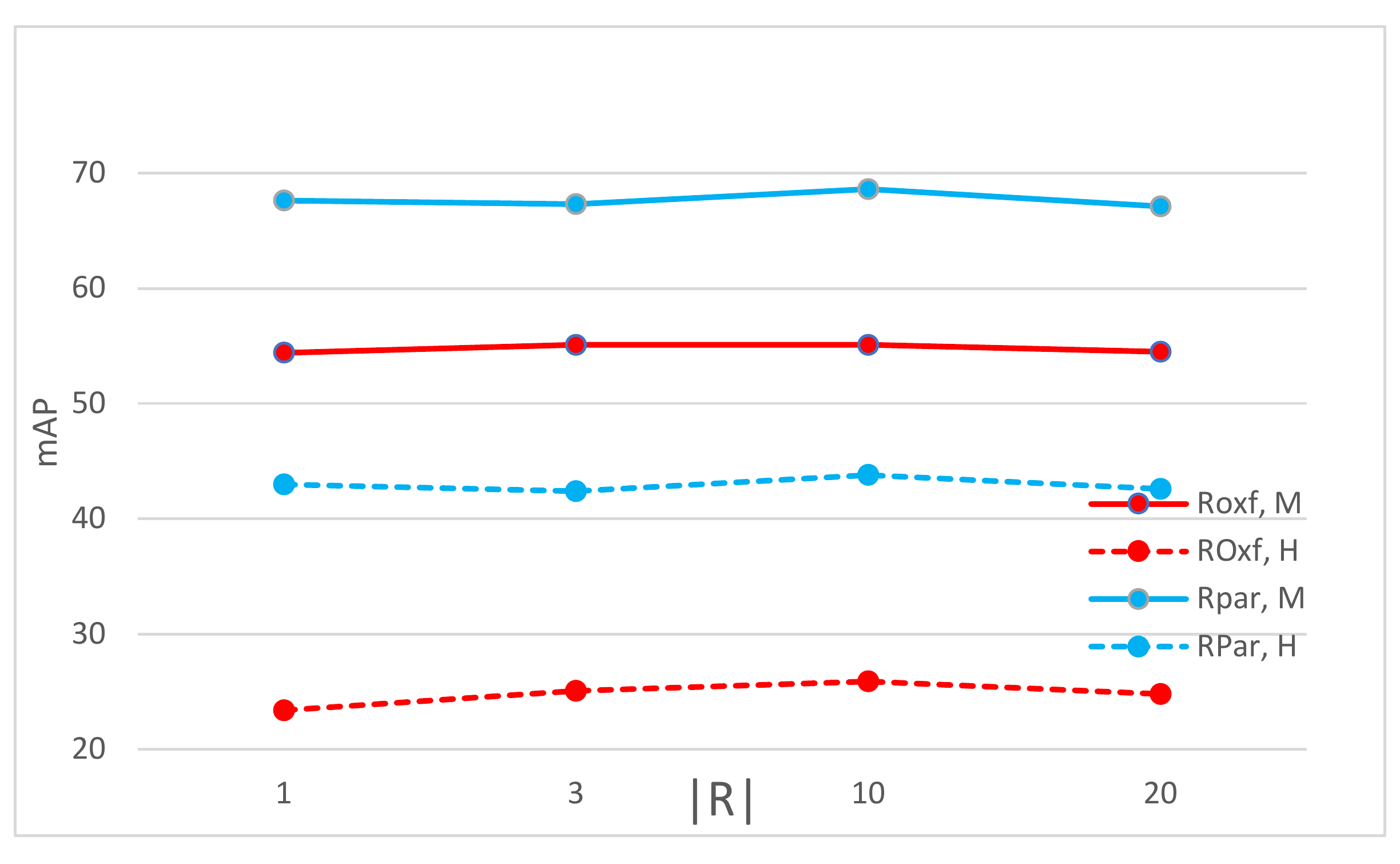}
  \caption{1b}
  \label{fig:sfig2}
\end{subfigure}
\caption{Figure shows the impact of different hyper-parameters $\tau$ (a) and $R$ (b) in retrieval performance.}
\label{fig:fig}
\end{figure*}

  In Fig. \ref{fig:sfig1} we study the impact of size of similarity threshold $\tau$ on the retrieval performance. As mentioned earlier, $ \tau$ controls the amount of semantic information that is distilled from teacher onto student model. From Fig. \ref{fig:sfig1} we observe that retrieval performance increases as $\tau$ is decreased. As explained earlier, high values of $\tau$ selects easy positives in the positive set, $P$. As we decrease this threshold, the hardness of positives increases. However, decreasing $\tau$ too much will allow false positives to get included in $P$ which will be detrimental to the learning process. This is evident by the sharp decrease in retrieval performance for $\tau$ = 0.65.

  Finally in Fig. \ref{fig:sfig2} we study the impact of $R$. We notice a marginal but consistent improvement in retrieval performance across both datasets as $R$ is increased. In particular for the Hard (H) setting, the performance improves by 2-3 $\%$ as $R$ increases from 1 to 10. Beyond $R$ = 10 there is a marginal drop in performance.

 The above experiments are in line with our motivation to apply the given hyper-parameters and also shows that beyond certain values, our proposed methods are not sensitive to the choice of hyper-parameter values.

 \section{Training Ablation}
 \label{sec:tr_abl}
 In this section we study some of the key components in the training algorithm. Firstly, in Tab. \ref{tab:results_enc} we show the results for different encoding methods: GeM, and MAC. In our experiments, the teacher model uses GeM encoding method. Results demonstrate that our proposed algorithm has comparable performance across different encoding methods. 
 
 Secondly, in Tab. \ref{tab:results_grad} we study the impact of back-propagating gradients beyond the level of mixed global representations. Results show that our proposed method of not propagating gradients through the mixed representations results in superior retrieval performance. Back-propagating beyond mixed representations results in over-fitting.
 
 Thirdly, we analyze the scenario where the original teacher representations are sparesely conncected. As such the mixed samples are located in low density regions resulting in most samples having empty positive sets from similarity labels (SL) alone. In such settings we expect the mixup labeling (ML) to provide training signals that can drive the learning process. Results are shown in Tab. \ref{tab:results_ML}. For this setting, we sample 1000 images from the full dataset such that each sample has atmost 3 Euclidean Nearest Neighbors. Results demonstrate that the mixup labelling significantly improves the retrieval performance across all settings in ROxford5k dataset. On RParis6k, both methods have comparable performance. No ML setting is compared under different teacher similarity thresholds to demonstrate that simply decreasing $\tau$ to increase occupancy of positive set, $P$ does not lead to improvement in performance. In addition, our method also outperforms the baseline setting without the proposed augmentations across both datasets.


\begin{table}[t!]
\newcolumntype{L}[1]{>{\raggedright\let\newline\\\arraybackslash\hspace{0pt}}m{#1}}
\newcolumntype{C}[1]{>{\centering\let\newline\\\arraybackslash\hspace{0pt}}m{#1}}
\newcolumntype{R}[1]{>{\raggedleft\let\newline\\\arraybackslash\hspace{0pt}}m{#1}}
\def\arraystretch{1.1}
\newcommand\cw{0.7cm}
\footnotesize

\begin{center}
{%
\setlength{\tabcolsep}{0.2mm}
\begin{tabular}{|@{~}L{3.5cm}|C{\cw}|C{\cw}|C{\cw}|C{\cw}|}
    \hline
    \multirow{2}{*}{Method}  &  \multicolumn{2}{c|}{ROxf} &  \multicolumn{2}{c|}{RPar} \\
    \cline{2-3}\cline{4-5}
    & M & H & M & H \\
    \hline\hline
    
    ResNet34 (GeM)  &  55.1 & 25.9 &  68.6 & 43.8 \\
    
    ResNet34 (MAC)  & 55.0 & 25.4 & 67.1 & 42.2 \\
   
    \hline
    
\end{tabular}
\caption{mAP performance for different encoding methods, GeM, MAC for the student network. Teacher model uses GeM encoding method.}
\label{tab:results_enc}
}
\end{center}
\end{table}

\begin{table}[t!]
\newcolumntype{L}[1]{>{\raggedright\let\newline\\\arraybackslash\hspace{0pt}}m{#1}}
\newcolumntype{C}[1]{>{\centering\let\newline\\\arraybackslash\hspace{0pt}}m{#1}}
\newcolumntype{R}[1]{>{\raggedleft\let\newline\\\arraybackslash\hspace{0pt}}m{#1}}
\def\arraystretch{1.1}
\newcommand\cw{0.7cm}
\footnotesize

\begin{center}
{%
\setlength{\tabcolsep}{0.2mm}
\begin{tabular}{|@{~}L{3.5cm}|C{\cw}|C{\cw}|C{\cw}|C{\cw}|}
    \hline
    \multirow{2}{*}{Method}  &  \multicolumn{2}{c|}{ROxf} &  \multicolumn{2}{c|}{RPar} \\
    \cline{2-3}\cline{4-5}
    & M & H & M & H \\
    \hline\hline
    
    ResNet34 (all grad) & 49.7 & 21.8 & 65.5 & 37.6 \\
    
    ResNet34  &  54.3 & 25.2 &  66.4 & 40.5 \\

    \hline
    
\end{tabular}
\caption{mAP performance with and without gradient back-propagation through augmented representations. Note that for this experiment we used $D$ = 2000.}
\label{tab:results_grad}
}
\end{center}
\end{table}

\begin{table}[t]
\newcolumntype{L}[1]{>{\raggedright\let\newline\\\arraybackslash\hspace{0pt}}m{#1}}
\newcolumntype{C}[1]{>{\centering\let\newline\\\arraybackslash\hspace{0pt}}m{#1}}
\newcolumntype{R}[1]{>{\raggedleft\let\newline\\\arraybackslash\hspace{0pt}}m{#1}}
\def\arraystretch{1.1}
\newcommand\cw{0.7cm}
\footnotesize

\begin{center}
{%
\setlength{\tabcolsep}{0.2mm}
\begin{tabular}{|@{~}L{4.0cm}|C{\cw}|C{\cw}|C{\cw}|C{\cw}|C{\cw}|C{\cw}|}
    \hline
    \multirow{2}{*}{Method}  &  \multicolumn{3}{c|}{ROxf} &  \multicolumn{3}{c|}{RPar} \\
    \cline{2-4}\cline{5-7}
    & E & M & H & E & M & H \\
    \hline\hline

    ResNet34 ($\tau=0.75$)  & 71.5 & 50.9 & 21.9 & 83.6 & 64.6 & 37.6 \\
    
    \hline
    
    ResNet34 (no-ML,$\tau=0.75$ )  & 66.7 & 47.2 & 18.5 & 82.8 & 63.9 & 37.8 \\

    ResNet34 (no-ML,$\tau=0.5$)  & 57.1 & 39.5 & 13.9 & 73.0 & 56.0 & 26.5 \\
    
    ResNet34 (no-ML,$\tau=0.65$)  & 64.2 & 46.3 & 17.7 & 78.0 & 59.5 & 30.4 \\
    
    ResNet34 (no-aug,$\tau=0.75$)  & 64.4 & 46.8 & 20.9 & 79.6 & 62.4 & 36.1 \\
    \hline
    
\end{tabular}
\caption{mAP performance with and without mixup labelling (ML). Note that for this experiment, $D$ was set to 1000.}
\label{tab:results_ML}
}
\end{center}
\end{table}

\section{Conclusion}
We have presented a knowledge distillation approach based on ranking distillation. The proposed approach transfers the ranking knowledge of a list of images from a cumbersome teacher onto a compact student model. The proposed method introduces key algorithmic design choices that make the approach data-efficient under budget constraints w.r.t access to black-box teacher model and the number of training samples.


Our results are comparable or better than the standard supervised methods with the same network architecture that are trained using full dataset. Under the training budget constraints, the proposed method clearly outperforms the baseline methods on challenging image retrieval datasets. Our approach finds use case in settings where teacher models are hosted as public APIs with limited access. 

\section*{Acknowledgement}
The authors would like to acknowledge Vikas Verma for providing helpful feedback in improving the manuscript. The computational resources provided by Aalto Science IT project and CSC servers, Finland is also acknowledged.

{\small
\bibliographystyle{ieee_fullname}
\bibliography{egbib}
}

\end{document}